\documentclass[times,twocolumn,final,authoryear]{elsarticle}

\usepackage{prletters}
\usepackage{framed,multirow}

\usepackage{amssymb}
\usepackage{latexsym}
\usepackage{lipsum}

\usepackage{url}
\usepackage{subfig}
\usepackage{float}
\usepackage{color}

\usepackage{enumitem,amssymb}
\newlist{todolist}{itemize}{2}
\setlist[todolist]{label=$\square$}
\usepackage{pifont}

\usepackage{multicol}
\usepackage{graphicx}
\usepackage{booktabs}
\usepackage{url}
\usepackage{xcolor}
\definecolor{newcolor}{rgb}{.8,.349,.1}

\usepackage{graphicx}
\usepackage{mwe}

\usepackage{color, colortbl}
\journal{Pattern Recognition Letters}

\begin{document}

\ifpreprint
  \setcounter{page}{1}
\else
  \setcounter{page}{1}
\fi

\begin{frontmatter}

% \title{An assessment on the exploitation of intrinsic image properties for tackling presentation attacks on facial recognition systems}

\title{FaceSpoof Buster: a Presentation Attack Detector Based on Intrinsic Image Properties and Deep Learning}

% \title{Exploiting intrinsic image properties for tackling presentation attacks on facial recognition systems}

\author[1]{Rodrigo \snm{Bresan}}
\author[2]{Allan \snm{Pinto}}
\author[2]{Anderson \snm{Rocha}}
\author[1]{Carlos \snm{Beluzo}}
\author[1,2]{Tiago \snm{Carvalho}}

\address[1]{Federal Institute of S\~{a}o Paulo (IFSP), Campinas, Brazil}
\address[2]{University of Campinas (UNICAMP), Campinas, Brazil}

\received{1 May 2013}
\finalform{10 May 2013}
\accepted{13 May 2013}
\availableonline{15 May 2013}
\communicated{S. Sarkar}

\begin{abstract}
Nowadays, the adoption of face recognition for biometric authentication systems is usual, mainly because this is one of the most accessible biometric modalities. Techniques that rely on trespassing these kind of systems by using a forged biometric sample, such as a printed paper or a recorded video of a genuine access, are known as presentation attacks, but may be also referred in the literature as face spoofing. Presentation attack detection is a crucial step for preventing this kind of unauthorized accesses into restricted areas and/or devices. In this paper, we propose a novel approach which relies in a combination between intrinsic image properties and deep neural networks to detect presentation attack attempts. Our method explores depth, salience and illumination maps, associated with a pre-trained Convolutional Neural Network in order to produce robust and discriminant features. Each one of these properties are individually classified and, in the end of the process, they are combined by a meta learning classifier, which achieves outstanding results on the most popular datasets for PAD. Results show that proposed method is able to overpass state-of-the-art results in an inter-dataset protocol, which is defined as the most challenging in the literature.
\end{abstract}


\end{frontmatter}

%\linenumbers

\section{Introduction}
% What is biometrics
The task of identifying a given individual by its physiological traits (e.g., face, iris or fingerprint) or behavioral patterns (e.g., keystroke dynamics, gait) is known as biometrics. Due to the major adoption of devices that rely on this kind of access, the development of new techniques that seek to impersonate a legitimate user increased significantly, adding up major challenges for security authentication systems. The process of attack a biometric system is known in the literature as presentation attack, but may also be referred as spoofing attack, and it consists in present to the acquisition sensor a synthetic biometric sample, containing the biometric pattern of a valid user, to authenticate itself as a legitimate user.

% Brief about this work 
In this work, we present a new tool named FaceSpoof Buster, for detecting presentation attacks, without the needs of any extra hardware components (e.g., depth sensor, infrared sensor).
% How
Using different intrinsic properties from a given biometric sample, the presented method reach great results, in comparison to previous works in the literature, with recognizable results on the task of classification on an unseen dataset, commonly known as inter-dataset evaluation.

Our hypothesis is based on the fact that, by extracting these intrinsic properties, such as depth, illumination, and saliency, we may obtain telltales that may reveal additional information about the authenticity from a given biometric sample.

Combining a Convolutional Neural Networks (CNN) and the transfer learning process, we are able to extract robust and discriminative features, which are combined with SVM classifiers in a two step classification process, to perform the detection of attack samples.  Our method outperforms many existing approaches proposed for face presentation attack detection (PAD) problem, with major emphasis on challenging tasks, such as the inter-dataset evaluation.

We can summarize our main contributions as follows: (1) proposition of a new method for face Presentation Attack Detection (PAD), named FaceSpoof Buster, which is based in a combination between intrinsic image properties and deep neural networks; (2) evaluation of different intrinsic properties (e.g., saliency, depth and illumination maps) for the problem of PAD, which to the best of our knowledge, have never been evaluated in this context; (3) an HTER that over overpass literature results in both inter and intra dataset protocol for different public datasets; (4) effective application of a previously trained CNN in a PAD context.

% A study of intrinsic properties in order to reveal cues that may help the task of Presentation Attack Detection;
%     \item An extendable framework for the adoption of new intrinsic features, possibly leading to better results in the task of PAD; 
%     \item A comparison between the proposed method and many others in tasks previously posed as challenging, such as inter-dataset protocol;
%     \item Results that outperform the current State-of-the-Art in previous datasets

\section{Related Works}
Accordingly to \cite{PAN_LIVENESS_DETECTION}, the techniques for presentation attack detection can be grouped into four major groups: user behavior modeling, data-driven characterization, user cooperation and hardware-based.

The techniques based on the first approach aims of recognizing presentation attacks by modeling the user's behavior such as head movements and eye blinking. Data-driven techniques are based on finding artifacts of attempted attacks by exploiting the data that came from a standard acquisition sensor. User cooperation based techniques seek to detect presentation attacks based on the interaction between the user and the authentication system, such as asking the user to execute some movement. Finally, there are techniques that use extra hardware, such as depth sensors and infrared cameras, in order to obtain more information about the scenario and thus be able to find cues that may reveal an attempted attack. Since this work focus on data-driven techniques, the rest of this section will be focused on this kind of methods.

\cite{SCHWARTZ_PARTIAL_LEAST_SQUARES} presented an anti-spoofing method by exploring the use of several visual descriptors for characterizing facial characteristics in terms of its color, texture, and shape properties. To deal with the high dimensionality of the final representation, the authors proposed the use of the Partial Least Squares (PLS) classifier, a statistical approach for dimensionality reduction and classification, which was designed to distinguish a genuine biometric sample from a fraudulent one.

\cite{PINTO_VISUAL_RHYTHM} proposed a data-driven method for video PAD based on Fourier analysis of the residual noise signature extracted from the input videos. The use of well-known texture feature descriptors, such as Local Binary Patterns was also considered in the literature by~\cite{MAATA_MICROTEXTURE}, which focuses on detecting micro-texture patterns that are added into the fake biometric samples during the acquisition process. 
Approaches based on Differences of Gaussian (DoG)~\citep{PEIXOTO_BAD_ILLUMINATION, TAN_SPARSE_LOW_RANK} and Histogram of Oriented Gradients (HOG)~\citep{KOMULAINEN_CONTEXT, YANG_COMPONENT_DESCRIPTOR} were also proposed, but at the cost of the results being affected by illumination conditions and the capture sensor, due to their nature.

\cite{YEH_MULTI_SCALE} proposed an effective approach againt face presentation attacks, based on perceptual image quality assessment, by adopting a Blind Image Quality Evaluatior (BIQE) along with a Effectivate Pixel Similary Deviation (EPSD), in order to generate new features to use on a multi-scale descriptor, showing it's efficacy when compared to previous works.

% \todo[inline]{Perhaps we should reformulate this last sentence since the Peixoto's method is kind of robust to bad illumination (maybe we could split this paragraph ...)}

% \todo[inline]{General comments: In my opinion, there are a lot of recent papers that aren't been discussed in this section. The most recent paper considered in this section date from 2013, which means this literature review is a little bit outdated. For example, in 2015, the T-IFS published a special issue for PAD, which contains several works that could be considered in this work.
% }

\section{Proposed Method}

% Distinguish between a genuine biometric sample and a synthetic one in scenarios with , as seen in previous works (\cite{PINTO_VISUAL_RHYTHM, YANG_CNN, PATEL_CROSS_DATABASE}). The problem relies on the fact there are many differences from the scenery of a given dataset to another, such as intrinsic properties (e.g., lighting, reflectance), as well as sensor characteristics (e.g., exposure, resolution).

% In order to tackle this problem, we propose a new method to perform the detection of presentation attacks, by using intrisinc image properties:

This section provide details about proposed method and each step of proposed framework. First step of proposed method will perform the frame extraction from the videos, followed by the extraction of intrinsic properties maps for each frame. Together, these maps represent specific properties (depth, illumination, and saliency) from a video along time.
% ~\todo[inline]{It would be great if we could put some rationales that justifies the use of such properties. \textbf{TODO}}

Then, we use a Convolutional Neural Network (CNN), specifically ResNet-50 (\cite{RESNET}), as a feature extractor in a way to encode properties from previously encoded maps. This process also takes advantage of a transfer learning (\cite{YOSINSKI_TRANSFER}) process, which uses previous ImageNet weights, since the number of samples to train our network from scratch is very restricted. Features extracted at this step are named bottleneck features.

\begin{figure*}[]
  \centering
  \includegraphics[width=0.98\textwidth]{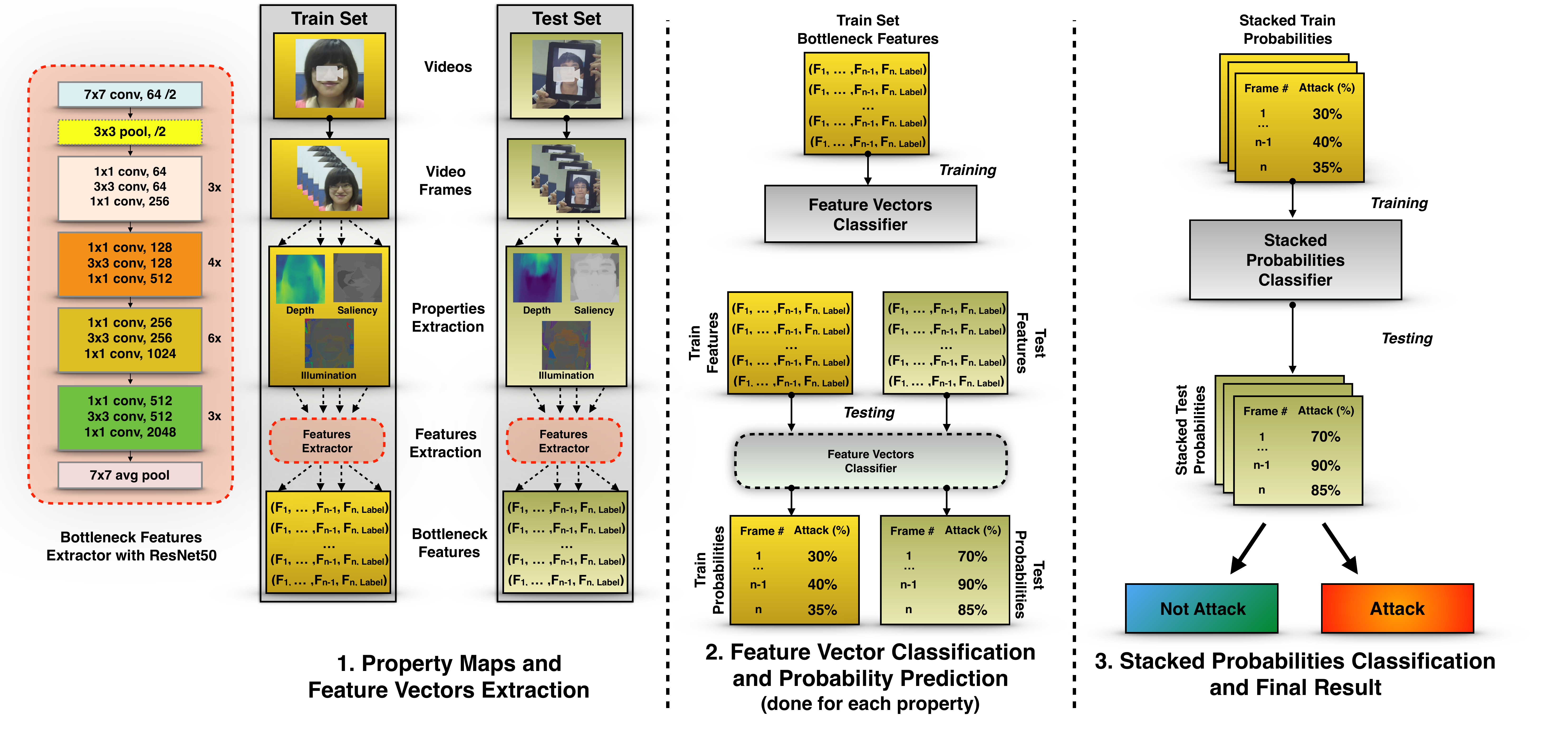}
      \caption{Overview of the proposed method. Initially, we represent intrinsic properties by using different kinds of maps, followed by the bottleneck feature extraction using the ResNet50 architecture; Followed by this, we perform the prediction of the previously extracted features, followed by the prediction of the probabilities; lastly, we perform the classification of the stacked probabilities, in order to generate the final predictions, to decide whether a given facial biometric sample is authentic or not.}
  \label{fig:overview}
\end{figure*}

% Next, we proceed to generate our first classifier, responsible for classifying a set of feature vectors, which will be from now on referred to as Feature Vectors classifier. For this task, we make usage of a Support Vector Machine (SVM) classifier with default parameters.
Once encoded, bottleneck features are classified using an SVM classifier, which provide confidence degree for each frame in an input video. These confidence scores are used in a final stage, which perform a meta-learning process, to train a new SVM classifier which combines information from illumination, depth and salience maps, resulting in a new artifact that will be referred to as probabilities vector. This new probabilities vector artifact is then fed to our second classifier, which is responsible for the final prediction for the tested samples. Fig.~\ref{fig:overview} depicts an overview of full proposed method.

\subsection{Frame extraction from videos}

Most of the benchmarks available for the face presentation attack detection problem are collections of videos, and for these datasets, we first need extracts the frames of each video, since the intrinsic property computation and the classification stage is performed upon images. 
% This task was accomplished by using the OpenCV library, which allows us to iterate over a single video, providing the retrieval of a frame from a given specified time. 
For this step, we perform a subsampling by extracting 10 frames per second from each video.
% \subsection{Properties Extraction}
% Proposed method relies over three different intrinsic properties: depth, saliency and illumination. Next sections provide details about estimation of each property.

\subsection{Depth Maps}

Due to the fact of presentation attacks being frequently reproduced over a flat surface, such as a sheet of paper with a printed face, or over a tablet reproducing a valid access, we believe that the depth estimation from a given biometric sample can provide relevant information about its authenticity, once that when presented with a flat surface, the estimated depth map should differ from real face.
% display a region with the same depth on all the points around the face.

% Many techniques for estimating depth information from a 2-D image have been proposed in the literature, such as proposed previously by \cite{FURUKAWA_MULTI} through the capture of several overlapping images from different viewpoints, as well as the work presented by \cite{DEPTH_PHOTOMETRIC} and \cite{DEPTH_HELIOMETRIC} by performing lighting changes on a static scene with fixed camera.

Our method estimates depth maps using the approach proposed by~\cite{MONODEPTH}, which uses stereo images and a fully convolutional  deep neural network associated with a modified loss function to estimates image depth. As in feature extraction step, described in Section~\ref{sec:bottleneckfeatures}, here we also take advantage of transfer learning methodology, transferring weights from~\cite{MONODEPTH} method's to our estimator.

% is based on an unsupervised approach for single depth estimation, showing great results even when compared with works that are based on supervised learning.

% By using a transfer learning approach, the work proposed by \cite{MONODEPTH} is based on an unsupervised approach for single depth estimation, showing great results even when compared with works that are based on supervised learning.

\subsection{Illumination Maps}

% In the work presented by \cite{TIAGO_ILLUMINATION}, the authors proposed a method for detecting image forgeries based on reflection properties from a single picture, once the forgeries will show differences in the reflection when compared to the original picture.

% We then proceed to make use of the method proposed by \cite{TIAGO_ILLUMINATION}, 
In digital forensics field, illumination inconsistencies have been constantly used to detect image forgeries (\cite{TIAGO_ILLUMINANT_BASED,TIAGO_ILLUMINATION}). The hypothesis of these works is that illumination is an important clue and very difficult to fake in authentication context. 

Inspired by~\cite{TIAGO_ILLUMINATION} approach, we also use illuminant maps to encode illumination information into PAD context. Our hypothesis is that generated illumination maps from a real face will show differences on its reflection when compared to the generated illumination map from a flat surface. This is similar to our hypothesis for the depth maps, but in this case, focusing on the fact that a real human face shows a surface that reflects in a different manner than a tablet or a sheet of paper.

Based on the work proposed by \cite{TAN_ILLUMINANT}, our method take advantage of Inverse Intensity-Chromaticity Space for estimating illuminant maps from a single image. 

% After a superpixel segmentation, we estimates illuminats for each superpixel using:

% %
% \begin{equation}
% \chi_c(\vec{x}) = m(\vec{x}) \frac{1}{\sum_{i\in\{R,G,B\}} f_i(\vec{x})} + \gamma_c,
% \end{equation}
% %
% where $\gamma_c$ denotes the chromaticity of the illuminant in channel $c$, whereas $m(\vec{x})$ mostly captures any geometric influences (i.e. light positioning, surface orientation and camera).

\subsection{Saliency Maps}

% Tasks related to the task of saliency map estimation from a single image have been shown promising results, as shown by \cite{ZHU_SALIENCY}. Our hypothesis is also based on the fact that the saliency maps from a given biometric sample may reveal additional cues that may help us in the task of identifying presentation attacks.

% \cite{ZHU_SALIENCY} proposed a novel method called boundary connectivity, allowing much higher results on the task of background estimation.  \cite{ZHU_SALIENCY} also proposed that, by using this estimated background, a novel method for generating the respective saliency maps, with the adoption of these previously generated background maps along with multiple low level cues. 
% The following equation denotes the method proposed by \cite{ZHU_SALIENCY}, in order to generate a saliency map from a given image.

% \begin{equation}
% B n d C o n ( R ) = \frac { | \{ p | p \in R , p \in B n d \} | } { \sqrt { | \{ p | p \in R \} | } }
% \end{equation}
% %
% where $p$ is a given image patch and \textit{Bnd} is the set of image boundary patches.

As in depth and illumination cases, our method also takes advantage of saliency information to encode valuable information into our method. Again, our hypothesis is that flat objects used in PAD will spoil quality in saliency estimation. 

Our saliency maps estimation is based on \cite{ZHU_SALIENCY} which have two major steps: (1) a background modeling using boundary connectivity, which characterizes the spatial layout of image regions with respect to
image boundaries; (2) a principled  optimization  framework  to  integrate
multiple low level cues, including proposed background measure.
\subsection{Bottleneck Features Extraction via ResNet50}
\label{sec:bottleneckfeatures}

Once our intrinsic properties are estimated, we perform an alignment at eye's level on all of our frames and their property maps, followed by a crop on the face region. The purpose of this extra step in to have all the frames normalized with the same alignment.
% , leading to a standardized database to feed our classifiers.

Then, our method takes advantage of transfer learning process~\cite{YOSINSKI_TRANSFER}, in a way to avoid the process of laborious handcraft feature extraction. We choose ResNet50 (\cite{RESNET}), a robust, well known and effective CNN architecture, associated with ImageNet weights, to extract features from previously generated maps. Removing top layer, we use ResNet50 architecture as a feature extractor, which provides feature vectors commonly known as bottleneck features.

As final output of this step, a feature vector of 2,048 dimensions will be generated, which we will be later on referred as bottleneck feature vector.
\subsection{Classification}

Adopting a two-step classification pipeline, in which the first classifier is used for feature vectors classification, while the latter one is used for classifying the probabilities generated by the first one, our method shows a major benefit when compared to previous works. By allowing us to stack together many intrinsic properties from a given frame, it's also possible to make use of additional information that may contribute to the task of PAD.

\subsubsection{Bottleneck Vectors Classifier}

For the task of bottleneck vectors classification, the adoption of a Support Vector Machine (SVM)~\citep{SVM} classifier was made, due to its robustness in the task of binary classification when using multiple features. Given a bottleneck feature vector, our classifier returns for each frame, the probability of that frame being an attack or not.

% class probabilities, once the usage of this probabilities will be used as input to our second classifier.

\subsubsection{Probability Feature Vector Assembly}

Given an input video $V_{P}$, which already have intrinsic properties estimated, composed by $n$ frames $f^{P}_{1}, f^{P}_{2},\cdots,f^{P}_{n}$, and where $I$ denotes the intrinsic property extracted from the video ($P \in \lbrace $D, I, S$ \rbrace$). In previous step (bottleneck vector classifier), we calculated the probability of each frame belonging to a class or another, denoted by $f^{P}_{i}$.

Using a fusion based approach, we combine information from all intrinsic image properties in a way to use all these information together, resulting in a probability feature vector ($pfv$) defined by
\begin{equation}
\centering
  pfv = \lbrace p^{D}, p^{I}, p^{S} \rbrace  
\end{equation}
where $p^{P}$ is given by
\begin{equation}
\centering
p^{P} = \frac{\sum_{i=1}^{n}f^{P}_{i}}{n}, \qquad P \in \lbrace D, I, S \rbrace
\end{equation}

% the results previously obtained from our previous feature vectors classifier, in order to generate a new artifact containing the probabilities predictions for all the property maps.

% This new artifact will be composed of the mean value of the probabilities from a given video, which was generated in the previous step.

\subsubsection{Probabilities Classifier}

As a final step of our classification pipeline, we proceed to feed our $pfv$ vectors into our second classifier, which will be later on referred as probabilities classifier. For this task, another SVM classifier was selected. As an output of this classification, we will have the final classification for each video.

% \begin{figure}[ht]
% \centering
% \subfloat[My first picture]{\label{fig:mdleft}{\includegraphics[width=0.4\textwidth]{figs/ra_illumination.png}}}\hfill
% \subfloat[My second picture]{\label{fig:mdright}{\includegraphics[width=0.4\textwidth]{figs/ra_illumination.png}}}
% \label{fig:subfigures}
% \end{figure}

\section{Experiments and Results}

For evaluation of the proposed method, different rounds of experiments were performed using three public anti-spoofing datasets, which contained samples from real accesses and from presentation attacks. The adoption of protocols focusing on intra-dataset evaluation, where one dataset is tested within the same scenario was performed by following the protocols suggested by datasets' creators. Evaluation between different datasets scenarios, commonly known as inter-dataset, was also conducted, in order to assess the performance of our method in unknown scenarios.

It is also paramount to realize that, since we are interested in evaluate the efficiency of each intrinsic property separately, final results reported for depth, illumination and saliency reflects a majority vote process among all the frames classified by the first classifier (without probabilities features classification).

% \begin{figure*}
%   \centering
%     \includegraphics[width=0.3\textwidth]{figs/cbsr_illumination.png}
%   \includegraphics[width=0.3\textwidth]{figs/ra_illumination.png}
%     \includegraphics[width=0.3\textwidth]{figs/nuaa_illumination.png}

%   \caption{TSNE Projection generated from the illumination maps for each dataset}
%   \label{fig:overview}
% \end{figure*}

% \begin{figure*}
%   \subfloat[CBSR]{\includegraphics[width=0.33\textwidth]{figs/cbsr_illumination.png}} 
%   \subfloat[Replay Attack]{\includegraphics[width=0.33\textwidth]{figs/ra_illumination.png}}
%   \subfloat[NUAA]{\includegraphics[width=0.33\textwidth]{figs/nuaa_illumination.png}}
%   \caption{Data visualization using t-SNE from the illumination maps for each selected dataset.}
%  \label{fig:tsne}
% \end{figure*}

\subsection{Datasets}

In order to address the efficiency of the proposed method, three publicly available anti-spoofing datasets were selected. The criteria for selection of these datasets among many others available was due to their major adoption in previous works that tackle PAD.

\subsubsection{Replay-Attack}

Consisting of 1300 video clips from both photo and video attacks from 50 subjects, the Replay-Attack dataset shows itself as a reliable dataset for the evaluation of our method, once it is presented with different lighting and environment conditions (\cite{REPLAY_ATTACK}). In this dataset, three different types of attack are provided: print attacks, reproduced by using a printed paper with the face of a legitimate user; mobile attacks, reproducing a valid user access over a mobile phone; and video attacks, similarly to the mobile attacks, but reproduced with a tablet. The Replay-Attack dataset is separated into three subsets: training set (containing 360 videos); development set (containing 360 videos); testing set (containing 480 videos); and enrollment set (containing 100 videos); 

\subsubsection{CASIA-FASD}

The CASIA-FASD dataset proposed by~\cite{CBSR} contains a total amount of 600 videos from 50 different subjects, created with the purpose of providing samples from many of the existent types of presentation attacks. The videos are presented in twelve different scenarios, where each of them is composed by three genuine accesses and three fraudulent from the same person. Three different resolutions were used to capture (low, normal and high), along with three different types of attack (normal, printed attacks, printed and warped, printed with cut on the eyes region and video-based attacks).

\subsubsection{NUAA Photograph Imposter Dataset}

The NUAA Photograph Imposter Dataset (\cite{TAN_SPARSE_LOW_RANK}) is composed of 15 subjects, comprising a total of 5,105 valid access images and 7,509 presentation attacks collected through a generic webcam at 20 fps with a resolution of 640 x 480 pixels. The subjects were captured over three sections in different places and lighting conditions. The production of the attack samples were made by shooting a high resolution photograph with a Canon digital camera. 

\subsection{Experimental Protocols}

In order to assess the performance of our method, we proceed the experiments by using two different protocols. The first one, consists in evaluates proposed method inside the same anti-spoofing dataset, which is commonly known as intra-dataset evaluation. The second one was conducted in order to address the efficacy of our method when tested on another dataset, commonly being referred as inter-dataset or cross-dataset evaluation. This later one is the most challenging in the literature, due to the differences in scenery that one dataset shows from another one.

For the intra-dataset evaluation, the authors dataset protocols were followed, in order to guarantee a better understanding of how our method performs when compared to other works.

As for the evaluation of our method on a previously unseen dataset, which is called inter-dataset, we followed the same protocols as defined by other previous works, where one database was used as train set, and another one was used as test set. For this evaluation, we also performed the combination between multiple datasets, in order to create a model that comprises characteristics from diverse scenarios and sensors, allowing us to generalize our model to better perform on new datasets.
\subsection{Experimental Setup}

The parameters configurations used in this work are described all over this section, in a way to provide an easy way to reproduce the results obtained by our method. 

Regarding the illumination maps and its segmentation, the used parameters are the same as the presented in the work of \cite{TIAGO_ILLUMINATION}.
% , with the max intensity being set to 0.9882, min intensity of 0.0588, $\sigma$ equals to 0.2, K equals to 300 and min size set to 15.
For the depth maps, the used parameters were also kept as default as proposed by \cite{MONODEPTH}.
% , by using the VGG 16 network, border as wrap mode and batch size set to 2.
In the task of generating the saliency maps, default values were also used without any changes.

We conducted the classification process by using two different classifiers. In the first one, our feature vectors classifier, an One-vs-Rest classifier was adopted using a Support Vector Machine (SVM) with default parameter values. The second classifier, used for predicting the class of a given set of probabilities, was performed by using also an SVM classifier, with the adoption of a Radial Basis Function (RBF) Kernel.
% , with gamma values of 1e-4 and 1e-3 and C values of 1, 10, 100, 1000 and 1000. For this second classifier, which was responsible to define if a given biometric sample was authentic or not, the input data was created by stacking the probabilities generated from the first classifier.

In this work, the major used frameworks were Keras (version 2.2.2) \footnote{https://keras.io} and TensorFlow (1.5.0)\footnote{https://www.tensorflow.org}\footnote{All the source code will be made freely available for usage and replication of the method hereby presented, upon paper acceptance.}.

\subsection{Intra-Dataset Evaluation}

In this section, the results obtained for the intra-dataset evaluation are presented, in order to evaluate how our method performs when testing within the same dataset. For the evaluation of our method, we used the same protocols defined by the databases authors, as well as the metrics proposed by them.

\subsubsection{CASIA}

% \begin{figure}[]
% \centering
%  \subfloat[short for lof][All]{
%   \includegraphics[width=0.50\linewidth]{figs/cbsr_all_roc.png}
%   \label{subfig:fig1}
%  }
%  \subfloat[short for lof][Cut]{
%   \includegraphics[width=0.50\linewidth]{figs/cbsr_cut_roc.png}
%   \label{subfig:fig2}
% }

%  \subfloat[short for lof][Print]{
%   \includegraphics[width=0.50\linewidth]{figs/cbsr_print_roc.png}
%   \label{subfig:fig3}
%  }
%  \subfloat[short for lof][Tablet]{
%   \includegraphics[width=0.50\linewidth]{figs/cbsr_tablet_roc.png}
%   \label{subfig:fig4}
% }
% \caption[short for lof]{ROC curves for Intra-Dataset Evaluation on the CASIA dataset for each attack type. \todo[inline]{I would replace the ROC curves by DET curves}}
% \label{fig:intra_casia}
% \end{figure}

In the evaluation of the CASIA dataset, the best overall result, in which all the attacks were evaluated together, was obtained by using the illumination maps property solely, achieving an HTER value of 3.88\%. The usage of the concatenated maps also showed great results for separated types of attack, such as when evaluated on the tablet and print attack types, achieving HTER values of 1.66\% and 3.88\%, respectively. Attacks performed by using a printed sheet of paper with a cut in the eyes region, the usage of the saliency maps showed great results, with a HTER value of 6.94\%.

Table~\ref{tab:intra_casia} provide all the obtained results for the intra-dataset evaluation on the CASIA-FASD database.

\begin{table}[!h]
  \centering
\caption{Performance Results (in \%) considering the Intra-Dataset Protocol for the CASIA dataset.}
\begin{tabular}{cccccccc} \toprule
  \textbf{Method} & \textbf{Tablet} & \textbf{Print} & \textbf{Cut} & \textbf{Overall} \\ \midrule
   
    Depth & 10.27 & 13.05 & 17.50 & 33.33 \\
    Illumination & 5.00 & 8.33 & 10.00 & \textbf{3.88} \\
    Saliency & 5.00 & 6.38 & \textbf{6.94} & 14.81 \\
    Concatenated & \textbf{1.66} & \textbf{3.88} & 11.38 & 15.55 \\\bottomrule

\end{tabular}
\label{tab:intra_casia}
\end{table}

% In the figure \ref{fig:intra_casia} all the ROC curves for each attack type and the used properties is presented. 
\subsubsection{Replay Attack Dataset}

As seen in the Table~\ref{tab:intra_ra}, the usage of the illumination maps showed a great performance on the evaluation of the Replay Attack dataset, with HTER values of 1.56\%, 1.25\%, 0.62\% and 5.50\% for High definition, Printed, Mobile and Overall attack types.

\begin{table}
  \centering
\caption{Performance Results (in \%) considering the Intra-Dataset Protocol for the Replay Attack dataset.}
\begin{tabular}{cccccccc} \toprule
  \textbf{Method} & \textbf{Highdef} & \textbf{Print} & \textbf{Mobile} & \textbf{All} \\ \midrule
   
    Depth & 30.93 & 16.87 & 22.81 & 31.62\\
    Illumination & \textbf{1.56} & \textbf{1.25} & \textbf{0.62} & \textbf{5.50} \\
    Saliency & 10.62 & 11.87 & 5.93 & 12.62 \\
    Concatenated & 10.00 & 7.50 & 10.00 & 10.75 \\\bottomrule
\end{tabular}
\label{tab:intra_ra}
\end{table}

\subsubsection{NUAA}

In Table~\ref{tab:intra_nuaa}, we display the results obtained for the intra-dataset evaluation protocol in the NUAA dataset. We can observe that the usage of depth maps in this dataset plays a major role on our performance results, achieving a HTER value of 20.75\% and a BPCER value of 3.15\%. 
\begin{table}
  \centering
  \caption{Performance Results (in \%) considering the Intra-Dataset Protocol for the NUAA dataset.} \label{tab:intra_nuaa}
\begin{tabular}{cccccccc} \toprule
  \textbf{Method} &\textbf{HTER} & \textbf{APCER} & \textbf{BPCER} \\ \midrule
   
    Depth & \textbf{20.75} & 38.36 & \textbf{3.15} \\
    Illumination & 48.20 & 29.37 & 67.04 \\
    Saliency & 34.20 & \textbf{27.89} & 41.11 \\
    Concatenated & 44.36 & 28.27 & 60.45 \\\bottomrule

\end{tabular}
\end{table}

\subsection{Inter-Dataset Evaluation}

Building a method that is highly adaptable from one face anti-spoofing database to another unknown one has been posed as a major challenge in previous works, and it's an essential ability for real world applications that rely on face recognition for authentication. This task is posed as a major challenge due to the differences in the scenery from a given database to another, such as illumination, depth, as well as hardware based configurations, such as capture sensor and camera processing.

In this section, we present the obtained results for the inter-dataset evaluation protocol, when one dataset was used for training and another was used for test. 

\subsubsection{CASIA}

Using a fusion based approach, we were able to achieve state of the art results for the CASIA dataset. By a combination of two different anti-spoofing databases (NUAA and Replay Attack), we obtained an HTER value of 33.14\%, as well as great results with the classification of the intrinsic properties alone, with HTER values of 36.82\% and 39.32\% for the depth and illumination maps, respectively.

\begin{table}[!h]
  \centering
  \caption{Performance Results (in \%) considering the Inter-Dataset Protocol for the CASIA dataset.} 
\begin{tabular}{cccccccc} \toprule
  \textbf{ Train Set} & \textbf{Method} & \textbf{HTER} & \textbf{APCER} & \textbf{BPCER} \\ \midrule
   
    \multirow{4}{*}{NUAA}  & Depth & 44.56 & 62.45 & 26.66 \\
                          & Illumination  & 55.34 & 84.01  & 26.66  \\
                          & Saliency  & \textbf{43.15} & 27.88  & 58.42 \\
                          & Concatenated  & 50.74 & 87.00  & 14.44      \\ \midrule
                          
    \multirow{4}{*}{\parbox{1cm}{Replay Attack}}  & Depth   & \textbf{43.33} & 8.88   & 77.78  \\
                                  & Illumination  & 50.18 & 0.30  & 100   \\
                                  & Saliency  & 52.79 & 8.92  & 96.67   \\
                                  & Concatenated  & 50.18 & 1.48  & 98.88      \\ \midrule
                                  
    \multirow{2}{*}{NUAA}           & Depth  & 36.82 & 31.59  & 42.04  \\
    \multirow{1}{*}{}               & Illumination   & 39.32 & 18.65   & 60.00   \\
    \multirow{1}{*}{\parbox{1cm}{Replay Attack}}  & Saliency  & 61.84 & 37.17  & 86.51 \\ 
                                    & Concatenated  & \textbf{33.14} & 46.29  & 20.00 \\\bottomrule

\end{tabular}
\label{tab:sometab}
\end{table}

\subsubsection{Replay Attack}

On the evaluation of the Replay-Attack dataset, the best results were obtained when using a combination between the CASIA-FASD and the NUAA databases, by making usage of the illumination maps, resulting in a HTER value of 36.75\%. Good results were also achieved by using only the NUAA dataset as train set along with the illumination maps, resulting in a HTER value of 41.64\%.

\begin{table}
  \centering
  \caption{Performance Results (in \%) considering the Inter-Dataset Protocol for the Replay Attack dataset.}
\begin{tabular}{cccccccc} \toprule
   \textbf{ Train Set} & \textbf{Method} & \textbf{HTER} & \textbf{APCER} & \textbf{BPCER} \\ \midrule
   
    \multirow{4}{*}{NUAA}  & Depth & 48.00 & 43.85 & 52.14  \\
                          & Illumination  & \textbf{41.64} & 3.28 & 80.00   \\
                          & Saliency  & 47.78 & 62.71 & 32.85\\
                          & Concatenated  & 43.42 & 31.14 & 55.71 \\ \midrule
                          
    \multirow{4}{*}{\parbox{1cm}{CASIA}}  & Depth & \textbf{53.00} & 22.25 & 83.75    \\
                                                  & Illumination & 55.37 & 88.25 & 22.5   \\
                                                  & Saliency  & 61.25 & 91.25 & 31.25  \\
                                                  & Concatenated  & 60.35 & 80.00 & 40.71       \\ \midrule
                                  
        \multirow{2}{*}{NUAA}           & Depth  & 47.62 & 27.75 & 67.50  \\
        \multirow{1}{*}{}               & Illumination   & \textbf{36.75} & 51.00 & 22.50   \\
    \multirow{1}{*}{CASIA}  & Saliency & 54.75 & 74.50 & 35.00   \\ 
                                    & Concatenated  & 40.21 & 37.57 & 42.85  \\\bottomrule
\end{tabular}
\label{tab:sometab}
\end{table}

\subsubsection{NUAA}

On the task of classifying the NUAA dataset, the best results were obtained for the depth maps, with a HTER value of 37.27\%, showing that this property, besides not performing so well in the Replay-Attack and the CASIA databases, still shows great potential for revealing cues that may indicate a presentation attack.

The usage of illumination maps also showed great results when used as training sets the Replay-Attack dataset and a combination between Replay-Attack and CASIA, achieving HTER values of 51.13\% and 50.51\%, respectively.

\begin{table}[!h]
  \centering
  \caption{Performance Results (in \%) considering the Inter-Dataset Protocol for the NUAA dataset.}
\begin{tabular}{cccccccc} \toprule
   \textbf{ Train Set} & \textbf{Method} & \textbf{HTER} & \textbf{APCER} & \textbf{BPCER} \\ \midrule
   
    \multirow{4}{*}{CASIA}  & Depth & \textbf{37.27} & 47.56 & 26.97 \\
                          & Illumination  & 50.88 & 11.66 & 90.09  \\
                          & Saliency  & 47.67 & 41.07 & 54.26 \\
                          & Concatenated  & 44.29 & 25.38 & 63.19 \\ \midrule
                          
    \multirow{4}{*}{\parbox{1cm}{Replay Attack}}  & Depth   & 64.94 & 52.84 & 77.03   \\
                                                  & Illumination  & \textbf{51.13} & 3.39 & 98.86  \\
                                                  & Saliency  & 56.29 & 15.30 & 97.29  \\
                                                  & Concatenated  & 52.55 & 9.80 & 95.29       \\ \midrule
                                  
        \multirow{2}{*}{CBSR}           & Depth  & 51.00 & 47.18 & 54.81  \\
        \multirow{1}{*}{}               & Illumination   & \textbf{50.51} & 4.39 & 96.63   \\
    \multirow{1}{*}{\parbox{1cm}{Replay Attack}}  & Saliency  & 60.26 & 33.52 & 86.99  \\ 
                                    & Concatenated  & 50.61 & 17.12 & 84.10 \\\bottomrule
\end{tabular}
\label{tab:sometab}
\end{table}
\section{Comparison with State-of-the-art}

In table \ref{tab:comparison_inter}, we display the results obtained by using our approach, for both the property maps individually as well for the concatenated ones. Significant results were achieved on the task of inter-dataset classification, mostly on the CASIA-FASD dataset, achieving an HTER value of 33.14\% when trained on the combination between the NUAA and the Replay-Attack datasets, overcoming results obtained in previous works (\cite{PINTO_VISUAL_RHYTHM, YANG_CNN}). The usage of property maps alone also showed great results for the CASIA dataset, with HTER values of 36.82\% and 39.32\%, respectively.

Outstanding results were also achieved for the Replay-Attack dataset when trained on the combination of NUAA and CASIA databases, achieving HTER values of 36.75\% when using the illumination maps, achieving near state of the art results for this dataset.

\begin{table}
  \centering
  \caption{Comparison Among Existing Aproaches Considering the Inter-Dataset Evaluation Protocol.}
\begin{tabular}{cccccccc} \toprule
  \textbf{Method} & \textbf{CASIA} & \textbf{Replay-Attack} & \textbf{NUAA}  \\ \midrule
   
   \cite{YEH_MULTI_SCALE} & 39.00 & 38.10 & - \\
    \cite{PINTO_IRIS} & 47.16 & 49.72 & - \\
    \cite{YANG_CNN} & 42.04 & 41.36 & -  \\
    \cite{PATEL_CROSS_DATABASE} & - & \textbf{31.60} & -  \\
    \cite{TAN_SPARSE_LOW_RANK} & - & - & 45.85 \\
    \cite{PEIXOTO_BAD_ILLUMINATON} & - & - &  49.85 \\
    Depth & 36.82 & 47.62 & 37.27 \\
    Illumination & 39.32 & \textbf{36.75} & 50.51 \\
    Saliency & 43.15 & 47.78 & 47.67 \\
    Proposed Method & \textbf{33.14} & 40.21 & 50.61 \\\bottomrule

\end{tabular}
 \label{tab:comparison_inter}
\end{table}

% \begin{figure*}[htb!]
% \centering
%   \subfloat[Raw]{\includegraphics[width=0.20\textwidth]{figs/raw.png}} \hspace{0.2cm}
%   \subfloat[Illumination]{\includegraphics[width=0.20\textwidth]{figs/illumination.png}}\hspace{0.2cm}
%   \subfloat[Saliency]{\includegraphics[width=0.20\textwidth]{figs/saliency.png}}\hspace{0.2cm}
%   \subfloat[Depth]{\includegraphics[width=0.20\textwidth]{figs/depth.png}}
%   \caption{Visual Assessment from Generated Property Maps}
%  \label{fig:tsne}
% \end{figure*}

% \section{Visual Assessment}

% \lipsum[3-4]

\section{Conclusions and Future Works}

In this paper, we have proposed a method that, by using a two-step classification model, along with intrinsic image properties, such as depth, illumination and saliency, is able to learn features for the task of presentation attack detection. 

Evaluating our method in three different databases, we overpass state-of-the-art results achieving a HTER value of 33.14\% for the CASIA-FASD dataset for the inter-dataset evaluation, which addresses the efficacy of our method when compared to previous works in the literature. This result, to the best of our knowledge, is the best achieved for the CASIA dataset, setting our method as the state of the art for this specific dataset.

We believe that the finds provided by this paper, such as the efficacy of using image intrinsic properties, can lead to a better understanding on the development of new anti-spoofing methods, as well as to provide better in the development of new datasets.

For future works, we plan to make usage of other PAD datasets, once we were able to achieve great results when combining more than one dataset for evaluation on the inter-dataset protocol, leading to even better results which can help to address how to tackle facial presentation attacks.

\section{Acknowledgments}

We would like to thank  S\~{a}o Paulo Research Foundation (FAPESP)(\#2017/12631-6), to the National Council for Scientific and Technological Development - CNPq (\#423797/2016-6), and to NVIDIA for the donation of a TITAN XP GPU to be used on this research.

\nocite{*}

\bibliographystyle{model2-names}
\bibliography{refs}

%%% TSNE, curvas ROC, tabelas de comparação (método sem mapas concatenados, com mapas concatenados)

\end{document}